\documentclass[10pt,twocolumn,twoside]{IEEEtran}
\makeatletter
\long\def\@makecaption#1#2{\ifx\@captype\@IEEEtablestring%
\footnotesize\begin{center}{\normalfont\footnotesize #1}\\
{\normalfont\footnotesize\scshape #2}\end{center}%
\@IEEEtablecaptionsepspace
\else
\@IEEEfigurecaptionsepspace
\setbox\@tempboxa\hbox{\normalfont\footnotesize {#1.}~~ #2}%
\ifdim \wd\@tempboxa >\hsize%
\setbox\@tempboxa\hbox{\normalfont\footnotesize {#1.}~~ }%
\parbox[t]{\hsize}{\normalfont\footnotesize \noindent\unhbox\@tempboxa#2}%
\else
\hbox to\hsize{\normalfont\footnotesize\hfil\box\@tempboxa\hfil}\fi\fi}
\makeatother

\usepackage[pdftex]{graphicx}
\graphicspath{{./fig/}}
\DeclareGraphicsExtensions{.pdf,.jpeg,.png}

\ifCLASSINFOpdf
\else
\fi
\usepackage[cmex10]{amsmath}
\usepackage{amsfonts}
\usepackage{algorithm}
\usepackage{algorithmic}
\usepackage{url}

\begin{document}
%
\title{Clustering with Multi-Layer Graphs: A Spectral Perspective}
%
%
%

\author{Xiaowen~Dong,
        Pascal~Frossard,
        Pierre~Vandergheynst
        and~Nikolai Nefedov
\thanks{X. Dong and P. Frossard are with Signal Processing Laboratory (LTS4), Institute of Electrical Engineering, 
Ecole Polytechnique F\'{e}d\'{e}rale de Lausanne (EPFL), CH-1015 Lausanne, Switzerland (e-mail: \{xiaowen.dong, pascal.frossard\}@epfl.ch).}
\thanks{X. Dong and P. Vandergheynst are with Signal Processing Laboratory (LTS2), Institute of Electrical Engineering, 
Ecole Polytechnique F\'{e}d\'{e}rale de Lausanne (EPFL), CH-1015 Lausanne, Switzerland (e-mail: \{xiaowen.dong, pierre.vandergheynst\}@epfl.ch).}
\thanks{N.Nefedov is with Nokia Research Center (NRC), Lausanne, Switzerland (e-mail: nikolai.nefedov@nokia.com).}
\thanks{}}

\maketitle

\begin{abstract}
Observational data usually comes with a multimodal nature, which means that it can be naturally represented by a multi-layer graph whose layers share the same set of vertices (users) with different edges (pairwise relationships). In this paper, we address the problem of combining different layers of the multi-layer graph for improved clustering of the vertices compared to using layers independently. We propose two novel methods, which are based on joint matrix factorization and graph regularization framework respectively, to efficiently combine the spectrum of the multiple graph layers, namely the eigenvectors of the graph Laplacian matrices. In each case, the resulting combination, which we call a ``joint spectrum" of multiple graphs, is used for clustering the vertices. We evaluate our approaches by simulations with several real world social network datasets. Results demonstrate the superior or competitive performance of the proposed methods over state-of-the-art technique and common baseline methods, such as co-regularization \cite{Kumar10} and summation of information from individual graphs.
\end{abstract}


%
\IEEEpeerreviewmaketitle

\section{Introduction}
%
%
%
%
\IEEEPARstart{C}{lustering} on graph is a problem that has been studied extensively for years. In this task we are usually given a set of objects, as well as an adjacency matrix capturing the pairwise relationships between these objects. This adjacency matrix is either represented by an unweighted graph, where the weight of edges is always equal to one, or a weighted graph, where the weight of edges can take any real positive values. The goal is to find an assignment of the objects into several subsets, such that the ones in the same subset are similar in some sense. Due to the wide range of applications for this problem, numerous approaches have been proposed in literature, and we point the readers to \cite{Schaeffer07} for an extensive survey on this topic. 

\begin{figure*}[t]
	\begin{center}
		\begin{tabular}{cc}
			~\includegraphics[width=0.30\textwidth]{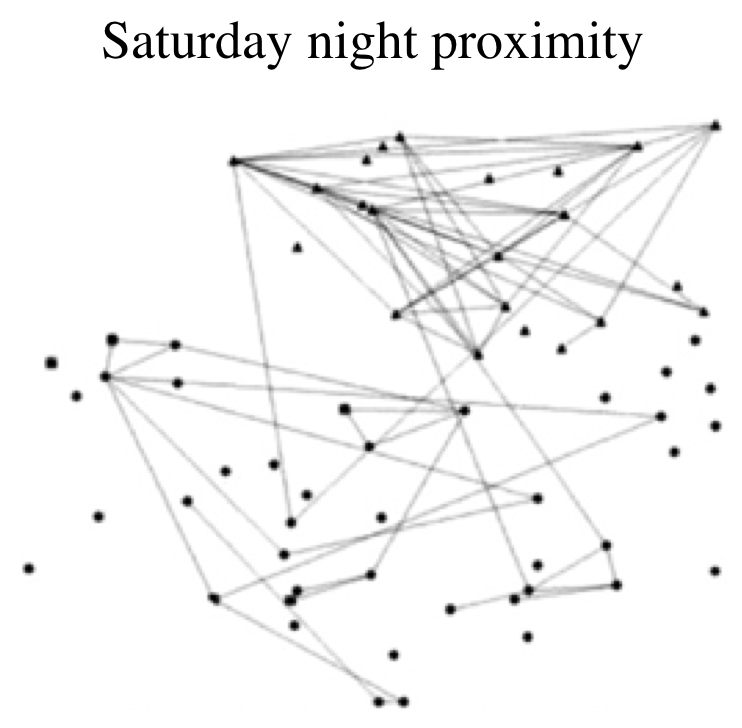}~ ~\includegraphics[width=0.30\textwidth]{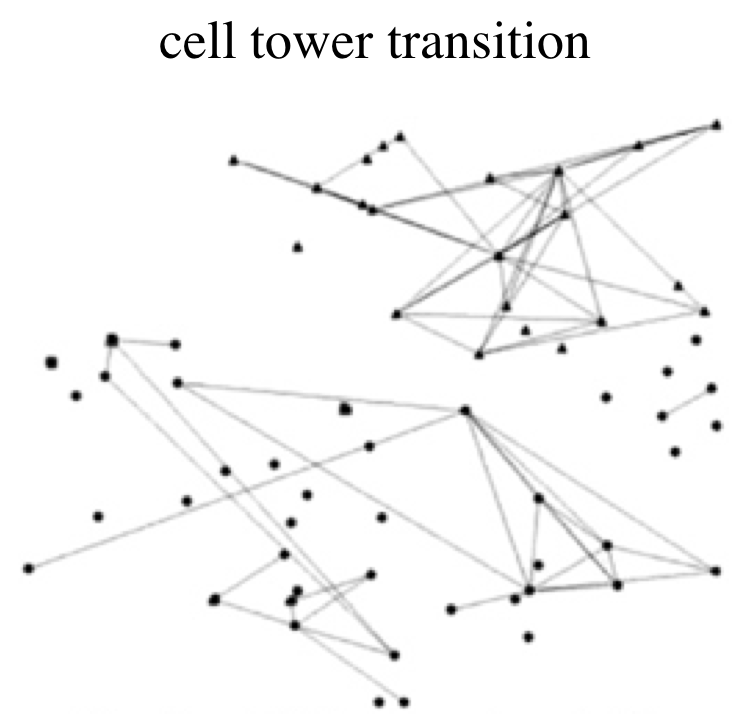}~ ~\includegraphics[width=0.30\textwidth]{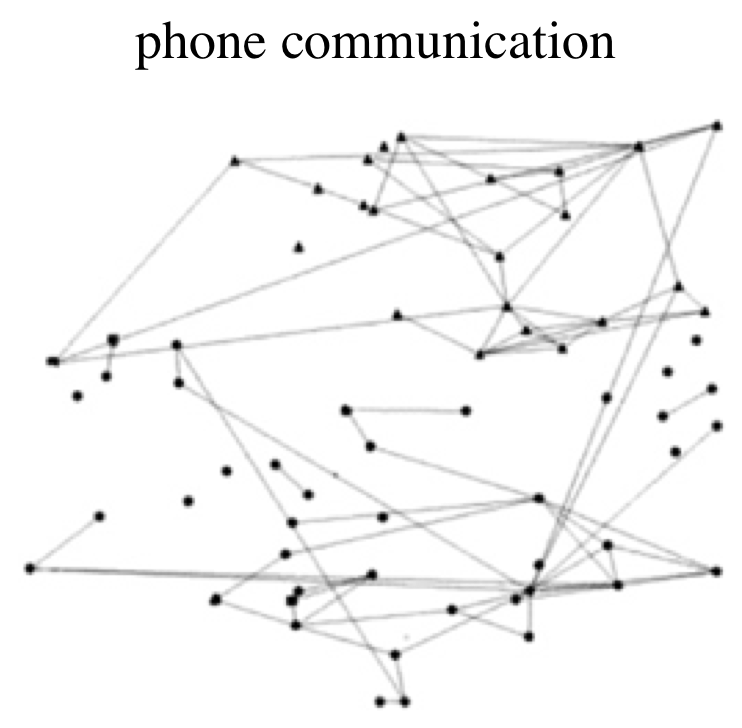}~
		\end{tabular}
	\end{center}
	\caption{A multi-layer graph in mobile social network \cite{Eagle10}: two mobile users are connected with an edge in the graph on the left if they are proximate to each other during a Saturday night; in the graph in the middle, two are linked together if they make the same cell tower transitions in the same time; on the right, we assign an edge between any pair who interacted with phone communication.}
	\label{fig:multigraph}
\end{figure*}

In contrast to the traditional problem, recent applications such as mobile and online social network analysis bring interesting new challenges. In these scenarios, it is common that observational data contains multiple modalities of information reflecting different aspects of human interactions. These different modalities can be conveniently represented by a multi-layer graph whose layers share the same set of vertices representing users, but have different sets of edges for each modality. Fig.~\ref{fig:multigraph} \cite{Eagle10} illustrates the mobile phone data collected in the MIT Reality Mining Project \cite{Eagle06} as such a multi-layer graph. Specifically, the graph layers represent relationships between mobile phone users in three different aspects: (i) Saturday night proximity, (ii) physical movement similarity and (iii) interaction with phone communication. Intuitively, each layer should contribute to a meaningful clustering result from its own angle; however, one can expect that a proper combination of the three graph layers will possibly lead to improved clustering results by efficient combination and completion of data in each layer.

\begin{figure*}[t]
	\begin{center}
		\begin{tabular}{cc}
			~\includegraphics[width=0.3\textwidth]{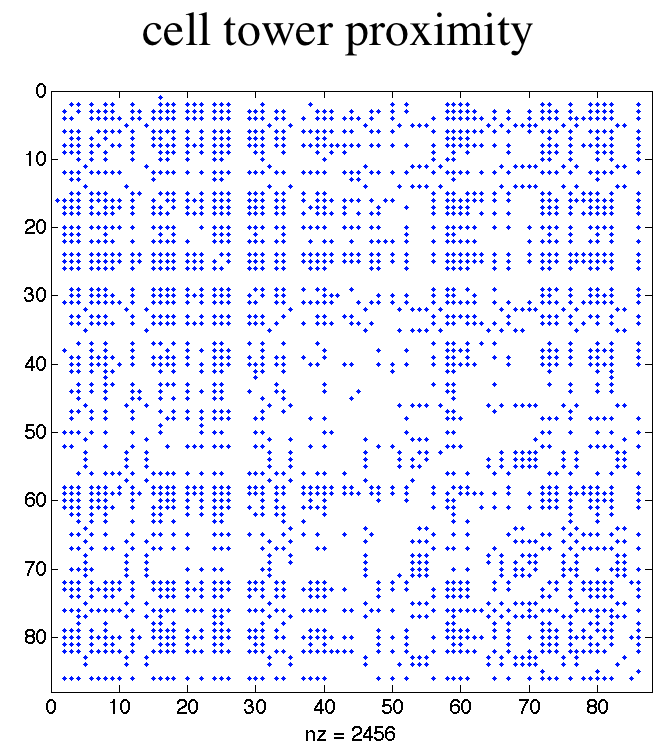}~ ~\includegraphics[width=0.3\textwidth]{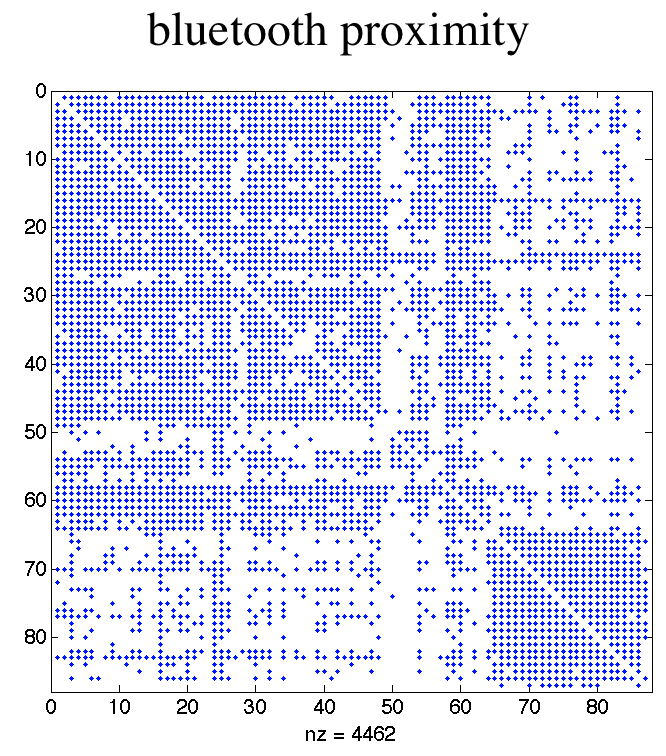}~ ~\includegraphics[width=0.3\textwidth]{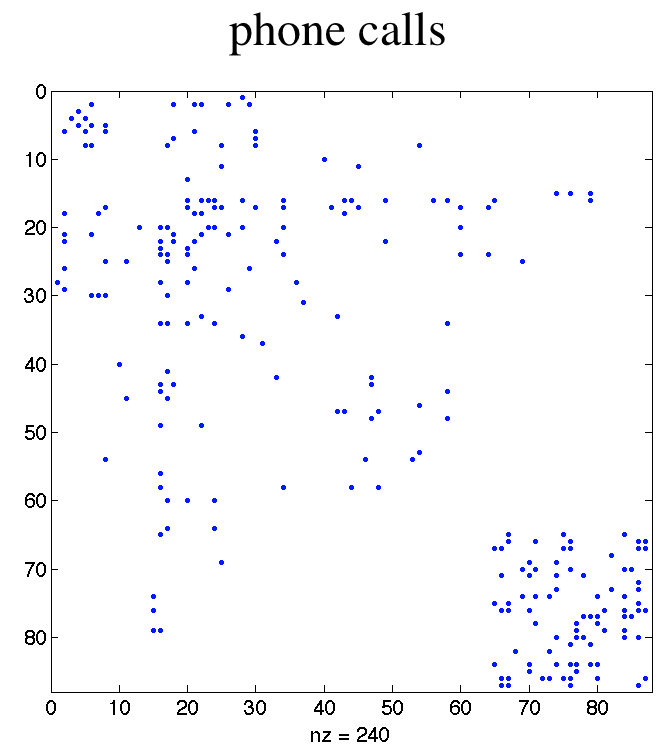}~\\
		\end{tabular}
	\end{center}
	\caption{Spy plots of three adjacency matrices from the MIT dataset: the redundant information contained in the cell tower and bluetooth proximities can compensate the sparse information from the phone calls for improved clustering results.}
	\label{fig:spy}
\end{figure*}

In this paper, we seek for such a good combination and propose two novel clustering methods by studying the spectrum of the graph. In particular, we propose efficient ways to combine spectrum of multiple graph layers, whose result is viewed as a ``joint spectrum" that is eventually used for spectral clustering \cite{Shi00}. In more details, we first propose to generalize the eigen-decomposition process applied on a single Laplacian matrix to the case of multiple graph Laplacian matrices. We design a joint matrix factorization framework in which each graph Laplacian is approximated by a set of joint eigenvectors shared by all the graph layers, as well as its specific eigenvalues from the eigen-decomposition. These joint eigenvectors can then be used to form a joint low dimensional embedding of the vertices in the graph, based on which we perform clustering. In a second approach, we propose a graph regularization method that combines the spectra of two graph layers. Specifically, we treat the eigenvectors of the Laplacian matrix from one graph as functions on the other graph. By enforcing the ``smoothness" of such functions on the graph through a novel regularization framework, we capture the characteristics of both graphs and get a better clustering result than with any graph alone. We finally propose an information-theoretic approach to generalize this second method to multiple graph layers.

We evaluate the performance of the proposed clustering methods on several real world social network datasets, and compare them with state-of-the-art technique as well as several baseline methods used for graph-based clustering, such as summation of information from individual graphs. The results show that, in terms of three clustering benchmarking metrics, our algorithms outperform the baseline methods, and are very competitive with the state-of-the-art technique introduced in \cite{Kumar10}. Furthermore, it is important to note that the contribution of this paper is not limited to a better clustering result with multiple graph layers. More generally, the concept of ``joint spectrum" is helpful to the analysis of multimodal data that can be conveniently modeled as a multi-layer graph. As an example, it can lead to the generalization of the classical spectral analysis framework to multi-dimensional cases.

The rest of the paper is organized as follows. In Section II, we formally introduce the problem of clustering with multi-layer graph and motivate it from a practical example. In Section III, we review briefly the spectral clustering algorithm, which is one of the building blocks of the methodologies proposed in this paper. Next, we describe in details our novel multi-layer clustering algorithms in Section IV and Section V. We then move onto simulations in Section VI, where we describe the datasets and present results and extensive comparisons with the existing methods. Finally, we list related work in Section VII and conclude the paper in Section VIII.

\section{Clustering with multi-layer graphs}
Consider a multi-layer graph $\mathcal{G}$\footnotemark[1] which contains $M$ individual graph layers $\mathcal{G}^{(i)}$, $i=1, \ldots, M$, where each layer $\mathcal{G}^{(i)}=\{V, E^{(i)}, \omega^{(i)}\}$ is a weighted and undirected graph consisting of a common vertex set $V$ and a specific edge set $E^{(i)}$ with associated weights $\omega^{(i)}$. Assuming that each layer reveals some aspect of the intrinsic relationships between the vertices, one can expect that a proper combination of information contained in the multiple graph layers possibly leads to improved unified clustering of the vertices in $V$. This can be further demonstrated by the following example.
\footnotetext[1]{Throughout the paper, the notation $\mathcal{G}$ without upper index still represents a single graph unless we explicitly mention that it is considered as a multi-layer graph.}
\footnotetext[2]{In these plots, the users are ordered according to 6 intended ``ground truth" clusters. However, one may find that it is not easy to distinguish the clusters from the observations, which in fact demonstrates the difficulty of this clustering task. Detailed discussions are in Section VI.}

\begin{figure*}[t]
	\begin{center}
		\begin{tabular}{cc}
			~\includegraphics[width=0.4\textwidth]{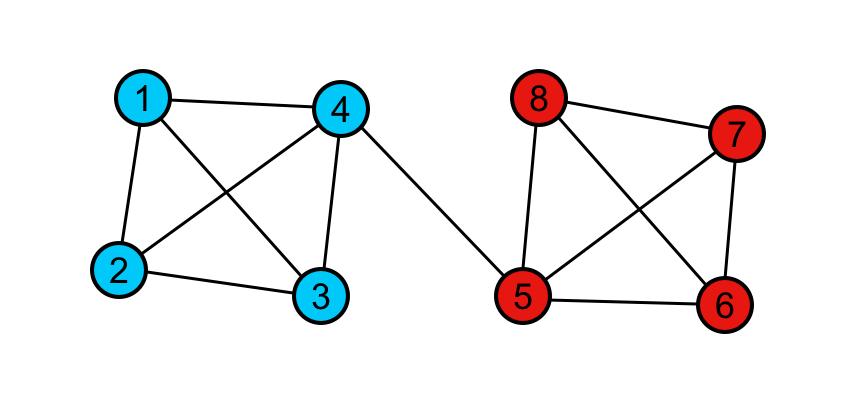}~ ~\includegraphics[width=0.4\textwidth]{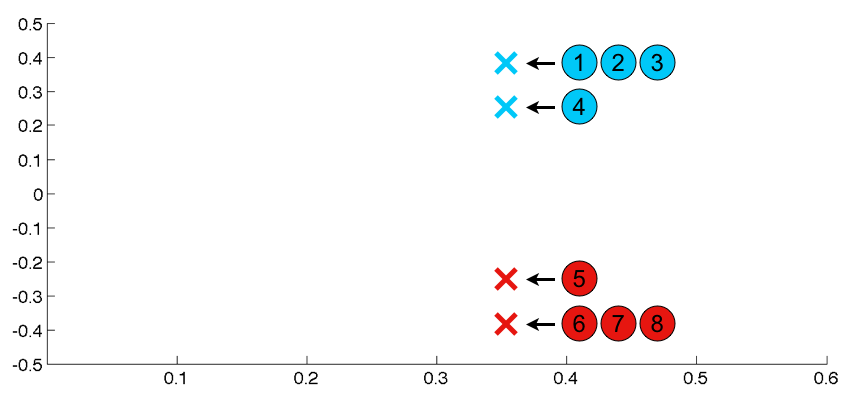}~\\
		\end{tabular}
	\end{center}
	\caption{Toy example to illustrate the spectral embedding. On the left is a simple unweighted graph with 8 vertices, which we want to partition into two clusters. On the right is the embedding of the original vertices into a 2-dimensional space using the spectrum of the graph: the coordinates on the horizontal and vertical axes are determined by the first and second eigenvectors of $L_{\text{rw}}$. In this case, vertices 1, 2 and 3 are embedded into the same point, and so are vertices 6, 7 and 8. It is clear to see that such an embedding helps reveal the intrinsic relationship between the vertices, and K-means can easily find the two clusters.}
	\label{fig:scexample}
\end{figure*}

Let us consider a three-layer graph built from the MIT Reality Mining Dataset \cite{Eagle09}, where vertices of the graph represent 87 participants of the MIT Reality Mining Project and edges represent relationships between these mobile phone users in terms of three different aspects, namely, cell tower proximity, bluetooth proximity and phone call relationship. From these graph layers we form three adjacency matrices and depict them in the spy plots in Fig.~\ref{fig:spy}, where each non-zero entry in the matrices corresponds to a point in the plots\footnotemark[2]. Intuitively, compared with the first two layers, entries in the phone call matrix are stronger indicators of friendship, hence the corresponding blue points in the third plot are more reliable. However, the sparse nature of this matrix makes it insufficient for achieving a good global clustering result of all the mobile users. In fact, this graph layer consists of many disconnected components, and it would be very difficult to assign cluster memberships to isolated vertices in the graph. In this case, the first two layers are more informative for achieving the clustering goal: even though each single entry there is less indicative, they provide richer structural information. This means that, by properly combining layers of different characteristics, we could expect a better unified clustering result.

In this paper, we address the following problem. Given a multi-layer graph $\mathcal{G}$ with $M$ individual layers $\mathcal{G}^{(i)}$, $i=1, \ldots, M$, we want to compute a joint spectrum that properly combines the information provided in different layers. In addition, the joint spectrum shall lead to an effective grouping of the vertices $V$ with spectral clustering \cite{Shi00}.

We propose two novel methods for the construction of a joint spectrum in the multi-layer graph.

\section{Spectral clustering}
The idea of working with the spectrum of the graph is inspired by the popular spectral clustering algorithm \cite{Shi00}. In this section, we give a very brief review of this algorithm applied on a single graph, which is the main building block of our novel clustering algorithms. Readers familiar with spectral clustering could skip this section.

Spectral clustering has become increasingly popular due to its simple implementation and promising performance in many graph-based clustering problems. It can be described as follows. Consider a weighted and undirected graph $\mathcal{G}$. The spectrum of $\mathcal{G}$ is represented by the eigenvalues and eigenvectors of the graph Laplacian matrix $L=D-W$ where $W$ is the adjacency matrix and $D$ is the degree matrix containing degrees of vertices along diagonal. Notice that $L$ is also called the unnormalized or combinatorial graph Laplacian matrix. There are two normalized versions of the graph Laplacian defined as follows:
\begin{eqnarray}
L_{\text{sym}}&=&D^{-\frac{1}{2}}(D-W)D^{-\frac{1}{2}} \\
L_{\text{rw}}&=&D^{-1}(D-W)
\end{eqnarray}
where $L_{\text{sym}}$ keeps the property of symmetry and $L_{\text{rw}}$ has close connection to random walk processes on graphs \cite{Luxburg07}. Different choices of the graph Laplacian correspond to different versions of the spectral clustering algorithm and detailed discussion on these choices is given in \cite{Luxburg07}. In this paper, we adopt the normalized spectral clustering algorithm that has been first described in \cite{Shi00}. It essentially corresponds to dealing with the eigenvalues and eigenvectors of the graph Laplacian $L_{\text{rw}}$. In practice, the algorithm finds the spectrum of $\mathcal{G}$, and embeds the original vertices in $\mathcal{G}$ to a low dimensional spectral domain formed by the graph spectrum. Due to the properties of the graph Laplacian matrix, this transformation enhances the intrinsic relationship among the original vertices. Consequently, clusters can be eventually detected in the new low dimensional space by many common clustering algorithms, such as the K-means algorithm \cite{MacQueen67}. An example of such an embedding is illustrated in the toy example shown in Fig.~\ref{fig:scexample}. An overview of the algorithm is given in Algorithm~\ref{alg:spectralclustering}.

\begin{algorithm}[h]
\caption{Normalized Spectral Clustering (\cite{Shi00})}

\begin{algorithmic}[1]

\STATE
\textbf{Input:} \\
$W$: The $n \times n$ weighted adjacency matrix of graph $\mathcal{G}$ with $n$ vertices\\
$k$: Target number of clusters\\

\STATE
Compute the degree matrix $D$.

\STATE
Compute the random walk graph Laplacian $L_{\text{rw}}=D^{-1}(D-W)$.

\STATE
Compute the first $k$ eigenvectors $u_1, \ldots, u_k$ (which correspond to the $k$ smallest eigenvalues)\footnotemark[3] of the eigenvalue problem $L_{\text{rw}}u=\lambda u$.

\STATE
Let $U\in \mathbb{R}^{n \times k}$ be the matrix containing $u_1, \ldots, u_k$ as columns.

\STATE
Let $y_i\in \mathbb{R}^k$ ($i = 1, \ldots, n$) be the $i$-th row of $U$ to represent the $i$-th vertex in the graph.

\STATE
Cluster $y_i$ in $\mathbb{R}^k$ into $C_1, \ldots, C_k$ using the K-means algorithm.

\STATE
\textbf{Output:} \\
$C_1, \ldots, C_k$: The cluster assignment\\

\end{algorithmic}
\label{alg:spectralclustering}
\end{algorithm}
\footnotetext[3]{Throughout the paper, eigenvalues and eigenvectors are always sorted in an ascending order, that is, $u_1$ is the eigenvector that corresponds to the smallest eigenvalue $\lambda_1$ and $u_n$ corresponds to the largest eigenvalue $\lambda_n$.}

As we can see in Algorithm~\ref{alg:spectralclustering}, the spectral embedding matrix $U$ consisting of the first $k$ eigenvectors of the graph Laplacian represents the key idea in spectral clustering. It gives a new representation $y_i$ for each vertex in this low dimensional space, which makes the clustering task trivial with the K-means algorithm. Moreover, as theoretical guarantees, \cite{Luxburg07} shows that the effectiveness of this approach can be explained from the viewpoint of several mathematical problems, such as the normalized graph-cut problem \cite{Shi00}, the random walk process on graphs \cite{Lovasz96} and problems in perturbation theory \cite{Stewart90}\cite{Bhatia97}. In the following two sections, we will generalize this idea to the case of multi-layer graphs, where we aim at finding a joint spectrum to form the spectral embedding matrix that represents information from all the graph layers.

\section{Clustering with generalized eigen-decomposition}
The first method that we propose for clustering with multi-layer graphs is built on the construction of an average spectral embedding matrix, based on which spectral clustering is eventually performed. We compute the average spectral embedding matrix with a generalized eigen-decomposition process. As we know, in order to compute the spectrum of a graph $\mathcal{G}$ with $n$ vertices, namely the eigenvalues and eigenvectors of its Laplacian matrix $L_{\text{rw}}$, one can compute an eigen-decomposition of the matrix $L_{\text{rw}}$ as:
\begin{equation}
L_{\text{rw}}=P\Lambda P^{(-1)}
\end{equation}
where $P$ is a $n \times n$ matrix containing eigenvectors of $L_{\text{rw}}$ as columns, and $\Lambda$ is a $n \times n$ diagonal matrix containing the corresponding eigenvalues as the diagonal entries. In case of a multi-layer graph $\mathcal{G}$ with $n$ vertices, we have $M$ Laplacian matrices ${L^{(i)}_{\text{rw}}}$, $i=1, \ldots, M$, one for each graph layer $\mathcal{G}^{(i)}$. As a natural extension, we propose to approximate each graph Laplacian ${L^{(i)}_{\text{rw}}}$ by a set of joint eigenvectors shared by all the graph layers as well as its specific eigenvalue matrix:
\begin{equation}
L^{(i)}_{\text{rw}}\approx P\Lambda^{(i)} P^{(-1)} \quad \mbox{for} \quad i=1, \ldots, M
\label{eq:approximation}
\end{equation}
where $P$ is a $n \times n$ matrix containing the set of joint eigenvectors as columns, and $\Lambda^{(i)}$ is the $n \times n$ eigenvalue matrix of $L^{(i)}_{\text{rw}}$. We now have to compute $P$, that is the set of eigenvectors that provides a good decomposition of the Laplacian matrix of all layers in the multi-layer graph. To do this, we propose to minimize the following objective function $S$, written as:
\begin{equation}
\begin{split}
\arg\min_{P, Q \in \mathbb{R}^{n \times n}} S=&\frac{1}{2}\sum_{i=1}^{M}||L^{(i)}_{\text{rw}}-P\Lambda^{(i)} Q||_F^2 \\
&+\frac{\alpha}{2}(||P||_F^2+||Q||_F^2)+\frac{\beta}{2}||PQ-I_n||_F^2
\end{split}
\label{eq:eigendecomp}
\end{equation}
where $P$ represents the joint eigenvectors, $Q$ is enforced to be the inverse matrix of $P$ so that it plays the role of $P^{(-1)}$ in Eq.~(\ref{eq:approximation}), and $\Lambda^{(i)}$ captures the characteristic of the $i$-th graph layer $\mathcal{G}^{(i)}$. In addition, $I_n$ represents the identity matrix of dimension $n$ and $||\cdot||_F$ denotes the Frobenius norm. Hence, the first term of the objective function $S$ is a data fidelity term to measure the overall approximation error when all layers are decomposed over $P$; the second term, the norms of $P$ and $Q$, are added to improve numerical stability of the solutions; and the third term is a constraint to enforce $Q$ to be the inverse of $P$. Notice that the purpose of introducing the additional variable $Q$ is mainly for the computational convenience of the optimization process. Finally, the regularization parameters $\alpha$ and $\beta$ balance the trade-off of the three terms in the objective function.

Now we have to solve the problem in Eq.~(\ref{eq:eigendecomp}) to get $P$. Since the objective $S$ is not jointly convex in $P$ and $Q$, it is difficult to find the global solution to Eq.~(\ref{eq:eigendecomp}). Therefore, we adopt an alternating scheme to find a local minimum of the objective function. In the outer loop, we first fix $Q$ and optimize $P$, and then optimize $Q$ while fixing $P$. As a consequence, it is important to give a good initialization to our algorithm. In practice, we suggest to compute the eigen-decomposition of $L_{\text{rw}}$ from the most informative graph layer, and initialize $P$ as the matrix containing its eigenvectors as columns. $Q$ is initialized as the inverse of $P$. The optimization process is then repeated until the stopping condition is satisfied. In the inner loop, we solve each variable while the other is being fixed. Notice that the objective function $S$ is differentiable with respect to variables $P$ and $Q$:
\begin{equation}
\frac{\partial S}{\partial P}=-\sum_{i=1}^{M}(L^{(i)}_{\text{rw}}-P\Lambda^{(i)} Q)Q^T\Lambda^{(i)}+\alpha P+\beta(PQ-I_n)Q^T
\end{equation}
\begin{equation}
\frac{\partial S}{\partial Q}=-\sum_{i=1}^{M}({L^{(i)}_{\text{rw}}}-P\Lambda^{(i)} Q)P\Lambda^{(i)}+\alpha Q+\beta(PQ-I_n)P
\end{equation}
Therefore we use an efficient quasi-Newton method (Limited-Memory BFGS \cite{Nocedal06}) to solve each variable.

We have now computed $P$, which is the set of joint eigenvectors, namely a joint spectrum shared by the multiple graph layers. The average spectral embedding matrix is then formed by the first $k$ joint eigenvectors, that is, the first $k$ columns of $P$. We then follow the steps 6 and 7 in Algorithm~\ref{alg:spectralclustering} to eventually perform the clustering. The updated algorithm is given in Algorithm~\ref{alg:eigen-decomposition}.

Notice that the algorithm proposed in this section is in a sense similar to \cite{Tang09}, which proposes a matrix factorization framework to find a low rank matrix that is shared by all the graph layers. However, the matrices they are trying to approximate are not the graph Laplacian matrices, but the adjacency matrices of all the layers. In addition, in their work, the approximation is done in a different way. Moreover, note that the generalized eigen-decomposition process above is essentially based on averaging the information from the multiple graph layers. It tends to treat each graph equally and to build a solution that smoothes out the specificities of each layer. In the next section, we propose a new method based on a regularization process between different layers, which is able to preserve the particularities of each individual layer.

\begin{algorithm}[h]
\caption{Clustering with generalized eigen-decomposition (\textbf{SC-GED})}

\begin{algorithmic}[1]

\STATE
\textbf{Input:} \\
$W^{(i)}$ ($i=1, \ldots, M$): $M$ $n \times n$ weighted adjacency matrices of a $M$-layer graph $\mathcal{G}$ with $n$ vertices\\
$k$: Target number of clusters\\

\STATE
For each $i$, compute the degree matrix $D^{(i)}$.

\STATE
For each $i$, compute the random walk graph Laplacian ${L^{(i)}_{\text{rw}}}={D^{(i)}}^{(-1)}(D^{(i)}-W^{(i)})$.

\STATE
Solve the optimization problem in Eq.~(\ref{eq:eigendecomp}) to get the joint eigenvector matrix $P$.

\STATE
Let $U'\in \mathbb{R}^{n \times k}$ be the matrix containing the first $k$ columns of $P$.

\STATE
Let $y_i\in \mathbb{R}^k$ ($i = 1, \ldots, n$) be the $i$-th row of $U'$ to represent the $i$-th vertex in the graph.

\STATE
Cluster $y_i$ in $\mathbb{R}^k$ into $C_1, \ldots, C_k$ using the K-means algorithm.

\STATE
\textbf{Output:} \\
$C_1, \ldots, C_k$: The cluster assignment\\

\end{algorithmic}
\label{alg:eigen-decomposition}
\end{algorithm}

\section{Clustering with spectral regularization}
In this section, we propose the second novel method for clustering with multiple graph layers, where we treat all layers based on their respective importance. As a consequence, this method helps preserve specificities of each layer in the clustering process.

\subsection{Intuition}
We first examine the behavior of eigenvectors of the graph Laplacian matrix in more details. Consider a weighted and connected graph $\mathcal{G}$ with vertex set $V=\{v_{i},i=1,\ldots,n\}$. From spectral graph theory \cite{Chung97}, we know that the eigenvectors $u_1, \ldots, u_n$ of the graph Laplacian matrix $L$ have the following properties:

\begin{enumerate}
\item The first eigenvalue $\lambda_1$ is 0 and the corresponding eigenvector $u_1$ is the constant one vector $\boldsymbol{1}$.
\item For $i = 2, \ldots, n$, $u_i$ satisfies: $u_i\perp \boldsymbol{1}$ and $||u_i||=1$ (after normalization).
\end{enumerate}

\begin{figure*}[t]
	\begin{center}
		\begin{tabular}{cc}
			~\includegraphics[width=0.3\textwidth]{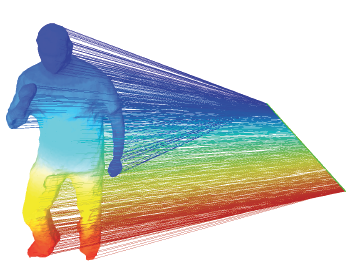}~&
			~\includegraphics[width=0.3\textwidth]{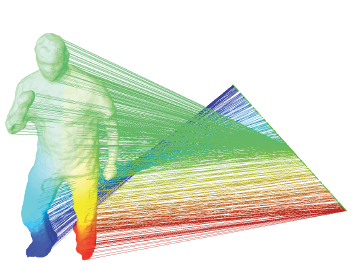}~\\
			~(a) mapping on $u_2$~ & ~(b) mapping on $u_3$~\\
			~\includegraphics[width=0.3\textwidth]{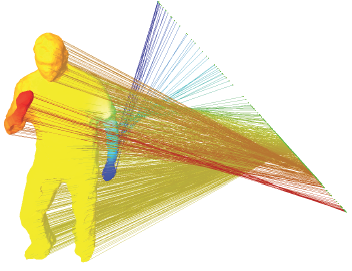}~&
			~\includegraphics[width=0.3\textwidth]{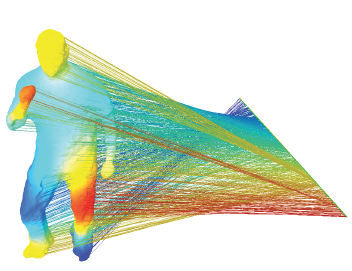}~\\
			~(c) mapping on $u_4$~ & ~(d) mapping on $u_8$~\\
		\end{tabular}
	\end{center}
	\caption{Examples of 1-dimensional mappings \cite{Horaud}.}
	\label{fig:mapping}
\end{figure*}

Now consider the problem of mapping the graph $\mathcal{G}$ on a 1-dimensional line such that connected vertices stay as close as possible on the line, while the mapping vector satisfies the second property above. In other words, we want to find a 1-dimensional mapping (or a scalar function) $f: V \rightarrow R$ that minimizes the following term:
\begin{equation}
\arg\min_{f \in \mathbb{R}^n} \Big\{\sum_{i,j}^n w_{i,j}(f(v_i)-f(v_j))^2\Big\}, \quad \mbox{s.t.} \quad f\perp \boldsymbol{1},~||f||=1.
\label{eq:quad1}
\end{equation}
where $f(v_i)$ and $f(v_j)$ represents the mapping of vertex $v_i$ and $v_j$ respectively, and $w_{i,j}$ is the weight of the edge between the two vertices. The constraints on the norm of $f$ and the orthogonality to the constant one vector $\boldsymbol{1}$ are introduced to make the solution nontrivial and unique, and can be explained from a graph-cut point of view \cite{Luxburg07}. Moreover, since eigenvectors of the Laplacian matrix can be viewed as scalar functions defined on the vertices of the graph, these conditions suggest that they can be considered as candidate solutions to the problem in Eq.~(\ref{eq:quad1}). In fact, we can rewrite Eq.~(\ref{eq:quad1}) in terms of the graph Laplacian matrix $L$ so that an equivalent problem is:
\begin{equation}
\arg\min_{f \in \mathbb{R}^n} f^{T}Lf, \quad \mbox{s.t.} \quad f\perp \boldsymbol{1},~||f||=1.
\label{eq:quad2}
\end{equation}
And it can be shown by the Rayleigh-Ritz theorem \cite{Luxburg07} that the solution to the problem in Eq.~(\ref{eq:quad2}) is $u_2$, the eigenvector that corresponds to the second smallest eigenvalue of $L$, which is usually called the Fiedler vector of the graph.

As an illustrative example of such a mapping, a weighted graph $\mathcal{G}$ constructed from a 3-dimensional point cloud and its mapping on to the Fiedler vector $u_2$ are shown in Fig.~\ref{fig:mapping}(a) \cite{Horaud}\cite{KSMH09}\cite{SHKV09a}. It can be seen that this mapping indeed keeps the strongly connected vertices as close as possible on the line. More importantly, it is shown in \cite{Zhou04} that the quadratic objective in Eq.~(\ref{eq:quad2}) can be viewed as a smoothness measure of a scalar function $f$ defined on the vertex set of a graph $\mathcal{G}$, that is, $f$ has similar values on the vertices that are strongly connected in the graph. Therefore, the fact that it minimizes this objective implies that the Fiedler vector $u_2$ is a smooth function on the graph. In fact, since we have 
\begin{equation}
u_i^{T}Lu_i=\lambda_i, \quad \mbox{for} \quad i = 2, \ldots, n
\end{equation}
all the first $k$ eigenvectors tend to be smooth on the graph $\mathcal{G}$ provided that the first $k$ eigenvalues are sufficiently small. This is illustrated in Fig.~\ref{fig:mapping}(b), (c), (d) for $u_3$, $u_4$ and $u_8$, and we can see that closely related points stay quite close on the mappings they represent. Since these first $k$ eigenvectors are used to form the low dimensional embedding $U$ in the spectral clustering algorithm, such smoothness properties imply that a special set of smooth functions on the graph, such as eigenvectors of the graph Laplacian matrix, can well represent the graph connectivity and hence help in the clustering process. 

This inspires us for combining information from multiple graph layers, with help of a set of joint eigenvectors that are smooth on all the layers, hence capture all their characteristics. However, instead of treating all the layers equally, we try to highlight the specificities of different layers. Therefore, we propose the following methodology. Consider two graph layers $\mathcal{G}^{(1)}$ and $\mathcal{G}^{(2)}$. From the smoothness analysis above, we observe that the eigenvectors of the Laplacian matrix from $\mathcal{G}^{(1)}$ are smooth functions on $\mathcal{G}^{(1)}$; in the meantime, since they can also be viewed as scalar functions on the vertex set of $\mathcal{G}^{(2)}$, we try to enforce their smoothness on $\mathcal{G}^{(2)}$ as well. This leads to a set of joint eigenvectors that are smooth on both graph layers, namely a jointly smooth spectrum shared by $\mathcal{G}^{(1)}$ and $\mathcal{G}^{(2)}$; this spectrum captures the characteristics of both layers.

\subsection{Jointly smooth spectrum computation}
We propose a spectral regularization process to compute a jointly smooth spectrum of two graph layers $\mathcal{G}^{(1)}$ and $\mathcal{G}^{(2)}$ by solving the following optimization problem:
\begin{equation}
\arg\min_{f_i\in{\mathbb{R}^n}} \Big\{\frac{1}{2}||f_i-u_i||_2^2+\lambda \cdot \Phi_{f_i}\Big\} \quad \mbox{for} \quad i=2, \ldots, k
\label{eq:specreg}
\end{equation}
where $f_i: V \rightarrow R$ is a scalar function on the graph, $u_i$ is the $i$-th eigenvector from $\mathcal{G}^{(1)}$, and $\Phi_{f_i}={f^{T}_i}{L^{(2)}_{\text{sym}}}{f_i} $ is a quadratic term\footnotemark[4] from $\mathcal{G}^{(2)}$ which measures the smoothness of ${f_i}$ on $\mathcal{G}^{(2)}$. In the problem in Eq.~(\ref{eq:specreg}), we seek for a scalar function $f_i$ such that it is not only close to the eigenvector $u_i$ that comes from $\mathcal{G}^{(1)}$, but also sufficiently smooth on $\mathcal{G}^{(2)}$ in terms of the quadratic smoothness measure. This promotes the smoothness property of our solution $f_i$ on both of the graphs, so that $f_i$ can be considered as a joint eigenvector of $\mathcal{G}^{(1)}$ and $\mathcal{G}^{(2)}$. The regularization parameter $\lambda$ is used to balance the trade-off between the data fidelity term and the regularization term in the objective function.
\footnotetext[4]{Since the smoothness analysis in Part A can be easily generalized from $L$ to $L_{\text{sym}}$, here we follow \cite{Zhou04} to use $L_{\text{sym}}$ instead of $L$ for a better implementation of the algorithm.}

It is shown that the problem in Eq.~(\ref{eq:specreg}) has a closed form solution \cite{Zhou04}:
\begin{equation}
f^\ast_i=\mu(L_{\text{sym}}+\mu I)^{-1}u_i
\end{equation}
where $\mu=\frac{1}{\lambda}$. Furthermore, notice that for each $u_i$ there is an associated optimization problem (except for $i=1$ since the first eigenvector is a constant vector), hence by solving all these problems we get a set of joint eigenvectors $f_i$, $i = 2, \ldots, n$. Therefore, they can be viewed as a jointly smooth spectrum of $\mathcal{G}^{(1)}$ and $\mathcal{G}^{(2)}$. The first $k$ joint eigenvectors can then be used to form a spectral embedding matrix, based on which we perform clustering. The overall clustering algorithm is summarized in Algorithm~\ref{alg:spectral-regularization}.

It is worth noting that $\mathcal{G}^{(1)}$ and $\mathcal{G}^{(2)}$ play different roles in our framework. Specifically, $\mathcal{G}^{(1)}$ is used for the eigen-decomposition process to get the eigenvectors, and $\mathcal{G}^{(2)}$ is used as the graph structure for the regularization process. It is natural to choose the more informative layer as $\mathcal{G}^{(1)}$. Moreover, we can generalize the above framework to graphs with more than two layers. Specifically, we propose to start with the most informative graph layer $\mathcal{G}^{(1)}$, and search for the next layer $\mathcal{G}^{(2)}$ that maximizes the mutual information between $\mathcal{G}^{(1)}$ and $\mathcal{G}^{(2)}$. More clearly, the mutual information between two graph layers is introduced by interpreting clustering from each individual layer as a discrete distribution of the cluster memberships of the vertices. Therefore, it can be calculated by measuring the mutual information shared by two distributions using Eq.~(\ref{eq:nmi}). Next, after having the combination of the first two layers, we can repeat the process by maximizing the mutual information between the current combination and the next selected layer, until we include all the graph layers in the end. This provides a greedy approach to compute a jointly smooth spectrum of multi-layer graphs.

\begin{algorithm}[h]
\caption{Clustering with spectral regularization (\textbf{SC-SR})}

\begin{algorithmic}[1]

\STATE
\textbf{Input:} \\
$W^{(i)}$ ($i=1, 2$): $n \times n$ weighted adjacency matrices of two graph layers $\mathcal{G}^{(1)}$ and $\mathcal{G}^{(2)}$\\
$k$: Target number of clusters\\

\STATE
For $\mathcal{G}^{(1)}$, compute the degree matrix $D^{(1)}$.

\STATE
Compute the random walk graph Laplacian ${L^{(1)}_{\text{rw}}}={D^{(1)}}^{(-1)}(D^{(1)}-W^{(1)})$.

\STATE
Compute the first $k$ eigenvectors $u_1, \ldots, u_k$ of ${L^{(1)}_{\text{rw}}}$.

\STATE
Let $U\in \mathbb{R}^{n \times k}$ be the matrix containing $u_1, \ldots, u_k$ as columns.

\STATE
For $i = 2, \ldots, n$, solve the spectral regularization problem in Eq.~(\ref{eq:specreg}) for each $u_i$ and replace it with the solution $f_i$ in $U$ to form the new low dimensional embedding $U''$.

\STATE
Let $y_i\in \mathbb{R}^k$ ($i = 1, \ldots, n$) be the $i$-th row of $U''$ to represent the $i$-th vertex in the graph.

\STATE
Cluster $y_i$ in $\mathbb{R}^k$ into $C_1, \ldots, C_k$ using the K-means algorithm.

\STATE
\textbf{Output:} \\
$C_1, \ldots, C_k$: The cluster assignment\\

\end{algorithmic}
\label{alg:spectral-regularization}
\end{algorithm}

\subsection{Discussion}
In addition to the intuition provided above, we further explain here why the spectral regularization process is considered as a good way of combining spectrum of two graph layers.

We first interpret the combination of multiple layers from the viewpoint of label propagation \cite{Zhou04-2}\cite{Zhu03}\cite{Bousquet04}\cite{Zhou04-3}, which is proven to be an effective approach for graph-based semi-supervised learning. In label propagation, one usually has a similarity graph whose vertices represent objects and edges reflect the pairwise relationship between them. We let the initial labels of the vertices propagate towards their neighboring vertices to make inference, based on the relationships between them and their neighbors. This is exactly what the spectral regularization process in Eq.~(\ref{eq:specreg}) does. More clearly, the optimization problem in Eq.~(\ref{eq:specreg}) can be solved through an iterative process, where in each iteration we have for every vertex $v \in V$:
\begin{equation}
(f_i(v))^{[n+1]} \leftarrow \alpha((I-{L^{(2)}_{\text{sym}}})f_i^{[n]})(v)+(1-\alpha)u_i(v)
\end{equation}
where $u_i$ represents the initial values on the vertices and $f_i^{[n]}$ represents the values of $f_i$ at iteration $n$ \cite{Zhou04}. The parameter $\alpha$ is defined as $\alpha=\frac{\lambda}{1+\lambda}$ while $\lambda$ is the regularization parameter in Eq.~(\ref{eq:specreg}). In other words, the value at each vertex is updated by a convex combination of the initial value $u_i(v)$ and the current values of its neighboring vertices, where the parameter $\alpha$ balances the trade-off between the two portions. Notice that the initial value $u_i$ from $\mathcal{G}^{(1)}$ is the continous-valued solver of a relaxed discrete graph-cut problem \cite{Luxburg07}. Therefore, $u_i$ can be viewed as labels indicating the cluster membership derived from $\mathcal{G}^{(1)}$. Consequently, the spectral regularization process in Eq.~(\ref{eq:specreg}) can be interpreted as a label propagation process, where the cluster labels derived from $\mathcal{G}^{(1)}$ are linearly propagated on $\mathcal{G}^{(2)}$. In this way, both of the graph structures have been taken into account hence making the resulting combination meaningful.

Another interpretation is based on disagreement minimization \cite{deSa05}\cite{Kumar10}, which has been proposed in the task of learning with multiple sources of data. The basic idea is to minimize the disagreement between information from the multiple sources so that we get a good representative of all the sources. For example, \cite{Kumar10} suggests a clustering algorithm that minimizes the disagreement between information from multiple graphs. Similarly, since we aim at finding a unified clustering result from multiple graph layers, it is natural to enforce the consistency between the clustering result and the information from all the graph layers, or in other words, to minimize the disagreement between them. Such a disagreement is again reflected in the objective function of the optimization problem in Eq.~(\ref{eq:specreg}). More specifically, the data fidelity term explicitly measures the disagreement between the solution $f_i$ and the initial value $u_i$ that comes from $\mathcal{G}^{(1)}$, while the regularization term implicitly represents the inconsistency of the information contained in $f_i$ with the structure of $\mathcal{G}^{(2)}$. Indeed, the regularization term $\Phi_f$ can be expressed in the following form:
\begin{equation}
\Phi_f=\frac{1}{2}\sum_{i,j}^n w_{i,j}\Big(\frac{f(v_i)}{\sqrt{d(v_i)}}-\frac{f(v_j)}{\sqrt{d(v_j)}}\Big)^2
\end{equation}
This means that $\Phi_{f_i}$ will only be small if the two end-point vertices of a large-weight edge in $\mathcal{G}^{(2)}$ have similar function values normalized by their degrees. Therefore, minimizing the objective function in Eq.~(\ref{eq:specreg}) can be considered as minimizing the total disagreement between the solution $f_i$ and the information from multiple graph layers. Notice that in this formation the disagreement is modeled from two different viewpoints for the two individual graphs, whose respective importance is controlled by the parameter $\lambda$.

\section{Simulation results}
In this section we present the experimental results. We first describe the datasets and different clustering algorithms used in the simulations, and then compare their performances in terms of three clustering benchmarking metrics.

\subsection{Datasets}
We adopt three real world social network datasets to compare the clustering performances between our proposed methods and the existing approaches. Two of them are mobile phone datasets, and the third one is a bibliographic dataset. In this section, we give a brief description on each dataset and explain how we construct multiple graph layers in each case.

The first dataset is the MIT Reality Mining Dataset, which includes mobile phone data of 87 mobile users on the MIT campus. We select three types of information to build the multi-layer graph: physical locations, bluetooth scans and phone calls. More specifically, for physical locations and bluetooth scans, we measure how many times two users are under the service of the same cell tower, and how many times two have scanned the same bluetooth device, within a 30-minute time window. Aggregating results from such windows throughout the 10-month period gives us two weighted adjacency matrices. In addition, a phone call matrix is generated by assigning weight of edge between any two users as how many times one has established or received calls from the other. In this dataset, we take the ground truth of clusters as the self-reported affiliations of the subjects, such as Media Lab graduate students and staff, and Sloan Business School students. The clustering goal is to partition all the users into 6 groups with the 3-layer graph and compare with the 6 intended clusters.

The second dataset we use is the mobile phone dataset that is currently being collected by Nokia Research Center (NRC) Lausanne in Switzerland \cite{Kiukkonen10}, which includes data of around 200 mobile users living or working in the area of Lausanne, Switzerland. We construct a multi-layer graph from the same information sources as that in the MIT dataset, with the only difference being that we measure the physical distance between every pair of users directly using their GPS coordinates. Therefore, this gives us a more accurate measure of the physical locations between these mobile users. In the Nokia dataset, we take the ground truth of clusters as 8 groups differentiated by their email affiliations reported in the questionnaire. The goal is to find the ground truth clusters with the multi-layer graph constructed.
\footnotetext[5]{Available online at ``\url{http://www.cs.umass.edu/~mccallum/data.html}" under category "Cora Research Paper Classification".}

The third dataset we adopt is the Cora dataset\footnotemark[5]. Although the objects of this bibliographic dataset are research papers rather than mobile users, it still reflects human interactions through research and publishing activities. In our experiments, we select 292 research papers that roughly come from three different communities: Natural Language Processing, Data Mining and Robotics. Each paper has been manually labeled with one of the categories and we consider this information as the ground truth of the clusters. To build the first two graph layers, we represent the title and abstract of each paper as vectors of nontrivial words, and take the cosine similarity between each pair as the corresponding entry in the adjacency matrix. In addition, we include a citation graph as the third layer that reflects the citation relationships of these papers. Finally, the goal is to cluster these papers based on the three graph layers we create.

It can be noted that the Cora dataset is considered quite easy to cluster while the MIT and Nokia datasets are much more difficult. The reason is that it is not straightforward to define the ground truth clusters between human users, and observational data does not necessarily correspond well to the intended clusters. In these two datasets, both the academic affiliations and email affiliations are not fully reflected by the physical proximity and phone communication between the mobile users, which makes the tasks difficult. Moreover, as we can imagine, the Nokia dataset is expected to be even more difficult than the MIT dataset as email affiliations is less trustworthy. Nevertheless, we still choose the ground truth clusters in this way as they are the best indicative information available in the datasets. After all, these two datasets are highly representative for analysis of rich mobile phone activities, and they can serve as challenging tasks in the evaluation compared with the easier one from the Cora dataset.

\subsection{Clustering algorithms}
In this section, we explain briefly the clustering algorithms that are included in the performance comparison, along with some implementation details. First of all, we describe some implementation details of the two proposes methods:
\begin{itemize}
\item \textbf{SC-GED}: Spectral Clustering with Generalized Eigen-Decomposition described in Section IV. In \textbf{SC-GED}, there are two regularization parameters $\alpha$ and $\beta$ to balance the approximation error and the stability and conditions on the solution. In our experiments, we set $\beta$ to be rather large, for example 100, to enforce the inverse relationship between $P$ and $Q$. We choose $\alpha$ to be 0.5 for the Nokia dataset and around 10 for the other two datasets.
\item \textbf{SC-SR}: Spectral Clustering with Spectral Regularization described in Section V. Since \textbf{SC-SR} is an recursive approach, we need to select two graph layers to fit the regularization framework at each time. As discussed in Section V Part B, we investigate the mutual information between different graph layers. As an example, in the MIT dataset, the ``cell tower" and ``bluetooth" layers have the highest mutual information. Therefore we choose to first combine these two layers. We select the ``bluetooth" layer to act as $\mathcal{G}^{(1)}$ in the spectral regularization framework, as it is considered more informative than the ``cell tower" layer. After the first combination, the third layer ``phone call" is incorporated to get the final solution. In addition, at each combination step, there is a regularization parameter $\lambda$ in the optimization problem in Eq.~\ref{eq:specreg} to control the relative importance of the two graph layers. Intuitively, the choice of this parameter at each step should loosely reflect the mutual information shared by the two layers being considered. We use this as a rule of thumb to set the parameters in the first and second combination step, which are denoted by $\lambda_1$ and $\lambda_2$, respectively. As an example, we set $\lambda_1=2$ and $\lambda_2=1$ in the MIT dataset.
\end{itemize}

Next, we introduce five competitor schemes as follows. The first three are common baseline methods for clustering with multiple graphs, and the other two are representative techniques in the literature:
\begin{itemize}
\item \textbf{SC-SUM}: Spectral clustering applied on the summation of adjacency matrices: 
\begin{equation}
\sum_{i=1}^M W^{(i)};
\end{equation}
If the weights of edges are of different scales across the multiple layers, we use the summation of the normalized adjacency matrices:
\begin{equation}
\sum_{i=1}^M {D^{(i)}}^{-\frac{1}{2}}W^{(i)}{D^{(i)}}^{-\frac{1}{2}};
\end{equation}
\item \textbf{K-Kmeans}: Kernel K-means applied on the summation of spectral kernels of the adjacency matrices: 
\begin{equation}
\sum_{i=1}^M K^{(i)} \quad \mbox{with} \quad K^{(i)}=\sum_{k=1}^{d}u^{(i)}_k{u^{(i)}_k}^T
\end{equation}
where $d\ll n$ (number of vertices) and $u^{(i)}_k$ represents the $k$-th eigenvector of the Laplacian $L^{(i)}_{\text{sym}}$ from $\mathcal{G}^{(i)}$.
\item \textbf{SC-AL}: Spectral Clustering applied on the averaged random walk graph Laplacian matrix: 
\begin{equation}
\frac{1}{M}\sum_{i=1}^M {L^{(i)}_{\text{rw}}}
\end{equation}
\item {Co-Regularization (\textbf{CoR})}: The co-regularization approach proposed in \cite{Kumar10} is the latest state-of-the-art technique aimed at combing information from multiple graphs. In this work, the authors proposed to enforce the similarity between information from two different graphs where the similarity is measured by a linear kernel. In our experiments, we generalize their approach to multiple graphs and tune the hyperparameter $\lambda$ in their work to achieve the best clustering performance.
\item {Community detection via modularity maximization (\textbf{CD})}: In addition to spectral-based clustering algorithms, modularity maximization is an approach proposed by Newman et al \cite{Newman02}\cite{Newman04}\cite{Newman042} for community detection. We adopt the algorithm described in \cite{Nefedov11}, which applies modularity maximization \cite{Newman042} using fast greedy search algorithm \cite{Blondel08}. 
It uses the summation of normalized adjacency matrices to combine information from different graph layers.
\end{itemize}

\subsection{Evaluation criteria and Results}
To quantitatively evaluate the clustering performance, we compare the clusters $\Omega=\{\omega_1, \ldots, \omega_k\}$ we have computed with the intended ground truth classes $C=\{c_1, \ldots, c_k\}$. We adopt \textit{Purity}, \textit{Normalized Mutual Information (NMI)} and \textit{Rand Index (RI)} \cite{Manning08} as three criteria to evaluate the clustering performance from different angles. More specifically, \textit{Purity} is defined as:
\begin{equation}
\textit{Purity}(\Omega,C)=\frac{1}{N}\sum_k \underset{j}{\mbox{max}} |\omega_k \cap c_j|
\end{equation}
where $N$ is the total number of objects, and $|\omega_k \cap c_j|$ denotes the number of objects in the intersection of $\omega_k$ and $c_j$. Next, \textit{NMI} is defined as:
\begin{equation}
\textit{NMI}(\Omega,C)=\frac{I(\Omega;C)}{[H(\Omega)+H(C)]/2}
\label{eq:nmi}
\end{equation}
where $I$ is the mutual information between clusters $\Omega$ and classes $C$ , while $H(\Omega)$ and $H(C)$ represent the respective entropy of clusters and classes. Finally, when interpreting the clustering result as a series of binary decisions on each pair of objects, \textit{RI} is defined as:
\begin{equation}
\textit{RI}(\Omega,C)=\frac{\textit{TP}+\textit{TN}}{\textit{TP}+\textit{FP}+\textit{FN}+\textit{TN}}
\end{equation}
where $\textit{TP}, \textit{TN}, \textit{FP}, \textit{FN}$ represent true positive, true negative, false positive and false negative decisions, respectively.

\begin{figure*}[h!tbp]
	\begin{center}
		\begin{tabular}{cc}
			~\includegraphics[width=0.90\textwidth]{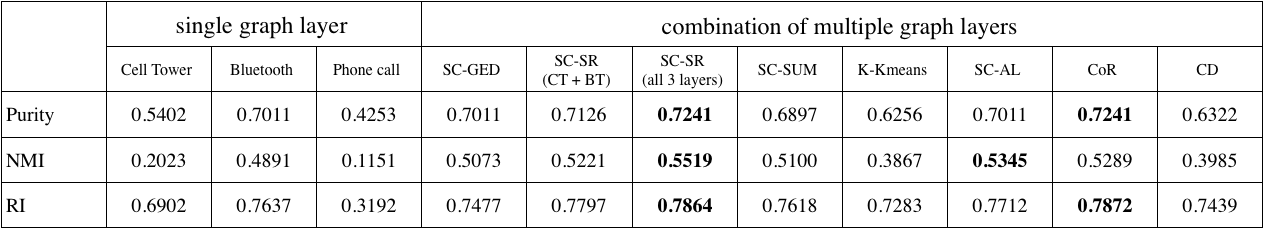}~ \vspace{0.3cm}\\
			\vspace{0.3cm}
			~(a) clustering performance on the MIT dataset~\\
			~\includegraphics[width=0.90\textwidth]{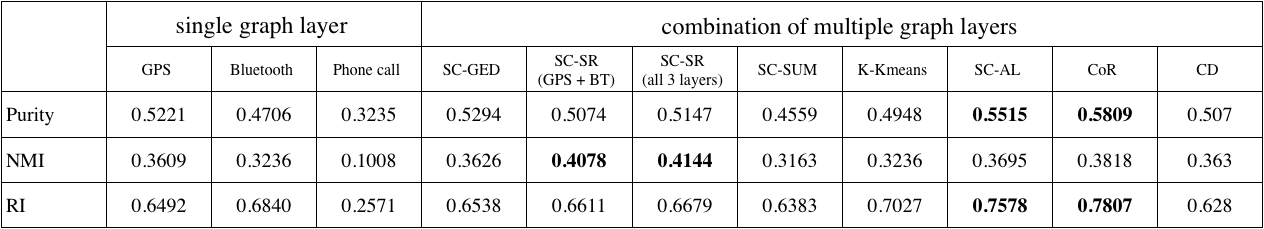}~ \vspace{0.3cm}\\
			\vspace{0.3cm}
			~(b) clustering performance on the Nokia dataset~\\
			~\includegraphics[width=0.90\textwidth]{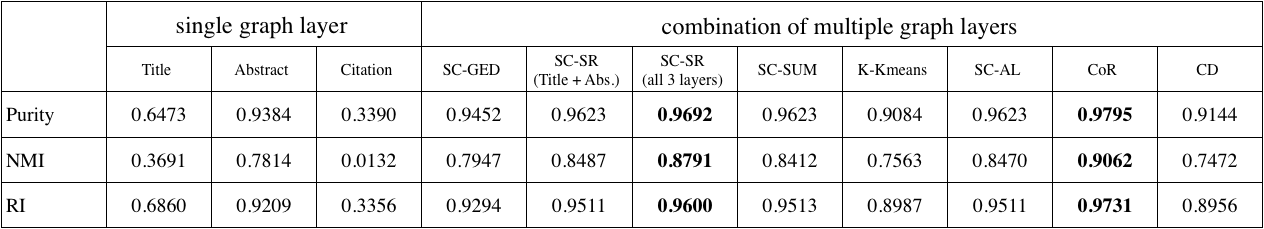}~ \vspace{0.3cm}\\
			\vspace{0.3cm}
			~(c) clustering performance on the Cora dataset~\\
		\end{tabular}
	\end{center}
	\caption{Performance evaluation of different clustering algorithms}
	\label{fig:result}
\end{figure*}

Fig.~\ref{fig:result} shows the performance for different clustering algorithms applied on the three datasets we adopt. For each scenario, the best two results are highlighted in bold font. As we can see, clustering with the Cora dataset is indeed much easier than the other two datasets as the benchmarks are much higher. Regarding the performance, it is clearly shown that proper combination of multiple graph layers indeed leads to improved clustering quality compared to using layers independently. In general, our proposed algorithm \textbf{SC-SR} achieves superior or competitive performance with the other combining methods in all the evaluation criteria, while \textbf{SC-GED} does not perform as well as \textbf{SC-SR}. Among the competitors, \textbf{CoR} presents impressive benchmarks, while \textbf{CD} and the baseline combining methods show intermediate results in general. As we can imagine, this is mainly due to the averaging of the information from different graph layers.

In more details, we can see that the regularized combination in \textbf{SC-SR} consistently leads to better benchmarking results as more layers are combined, particularly in terms of the \textit{NMI} scores. This comes from the way we combine the multiple graph layers, where the mutual information between them has been maximized. Compared to the state-of-the-art algorithm \textbf{CoR}, \textbf{SC-SR} maintains competitive results while the computational complexity is significantly reduced. Indeed, \textbf{CoR} needs to compute extremal eigenvectors of the (original and modified) Laplacian matrices for $n\frac{M(M-1)}{2}$ times in total, where $M$ is the number of different graphs and $n$ is the number of iterations the algorithm needs to converge. In contrast, \textbf{SC-SR} only needs to implement the same process once, namely for the most informative layer. Note that the performance in terms of \textit{NMI} shows differences with the other two criteria in the Nokia dataset, since the ground truth clusters in this dataset are quite unbalanced.

Compared with \textbf{SC-SR}, the performance of \textbf{SC-GED} is somehow disappointing, as it only provides limited improvement on the clustering quality achieved by individual layers. This is mainly due to the nature of the algorithm: unlike \textbf{SC-SR} which is implemented recursively, it resorts to a joint matrix factorization framework to find the set of joint eigenvectors all at once. Therefore, it can be essentially considered as a way to average the information from multiple sources, but without paying much attention to the specific characteristics they have. Nevertheless, we still believe that it is interesting in terms of the concept, and future work will be devoted to the improvement.

\begin{figure*}[h!tbp]
	\begin{center}
		\begin{tabular}{cc}
			~\includegraphics[width=0.22\textwidth]{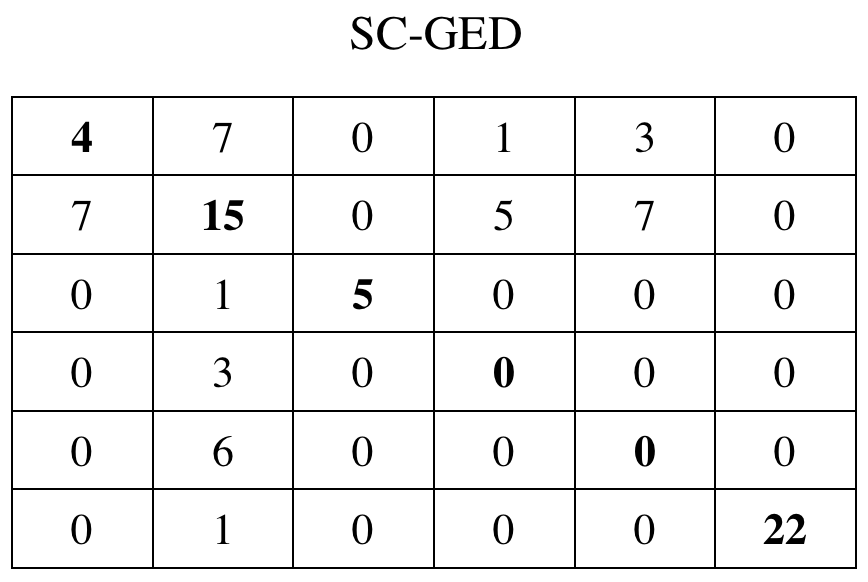}~ ~\includegraphics[width=0.22\textwidth]{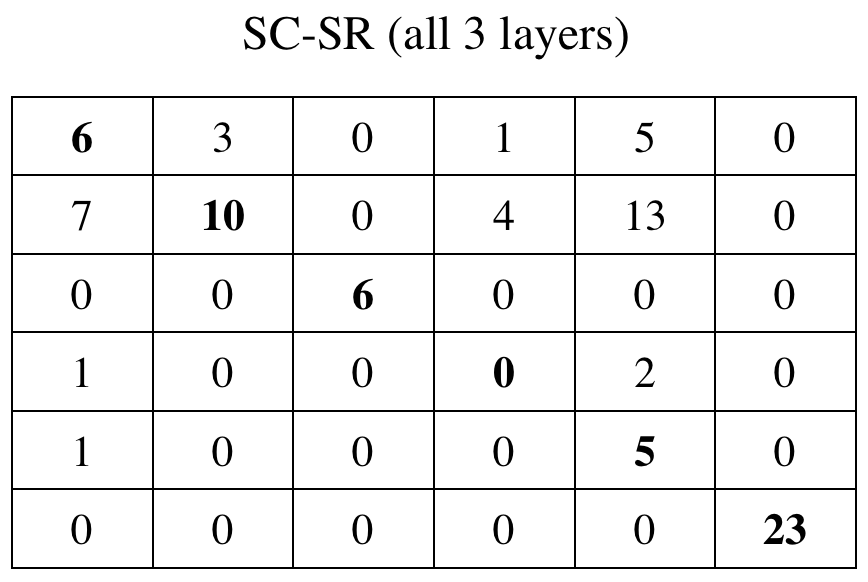}~ ~\includegraphics[width=0.22\textwidth]{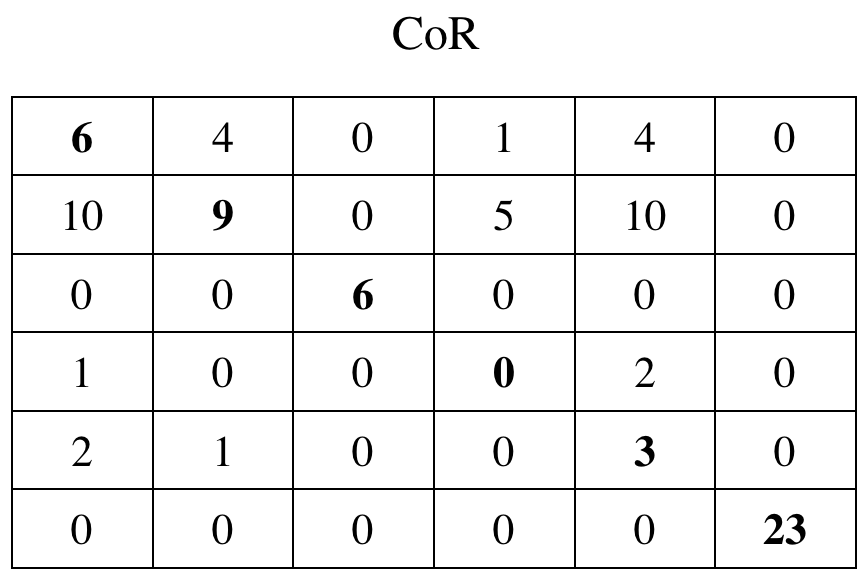}~ ~\includegraphics[width=0.22\textwidth]{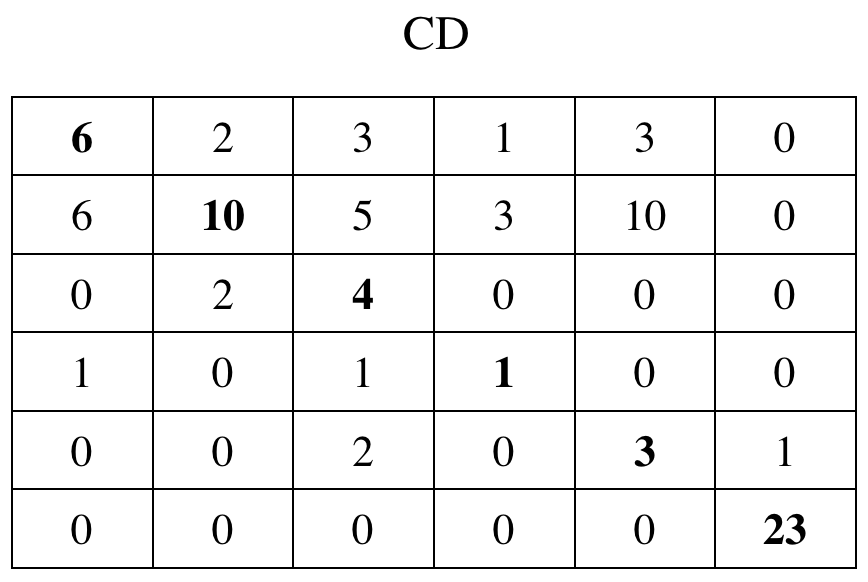}~\\
			~\includegraphics[width=0.22\textwidth]{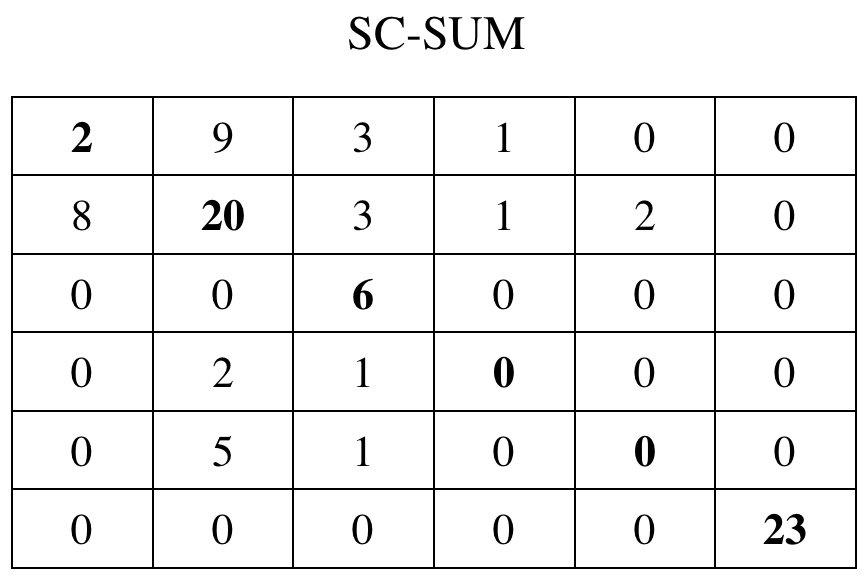}~ ~\includegraphics[width=0.22\textwidth]{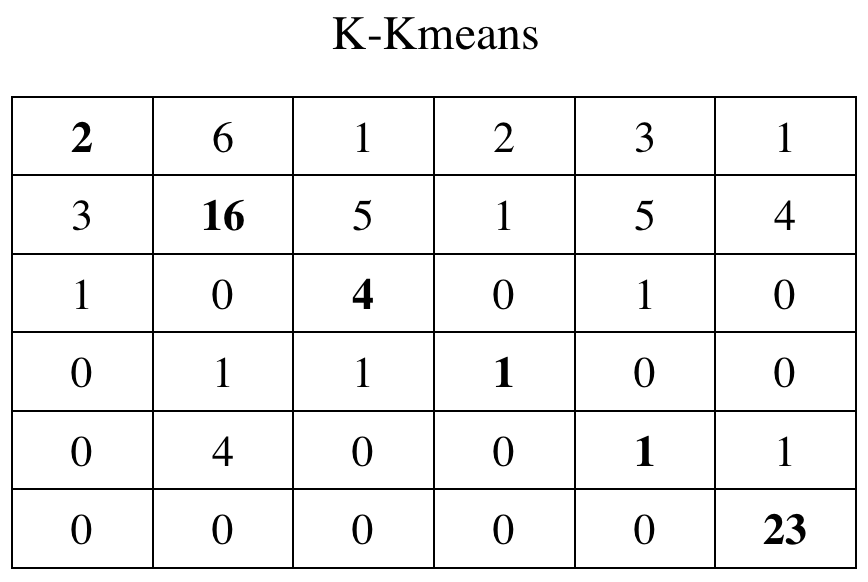}~ ~\includegraphics[width=0.22\textwidth]{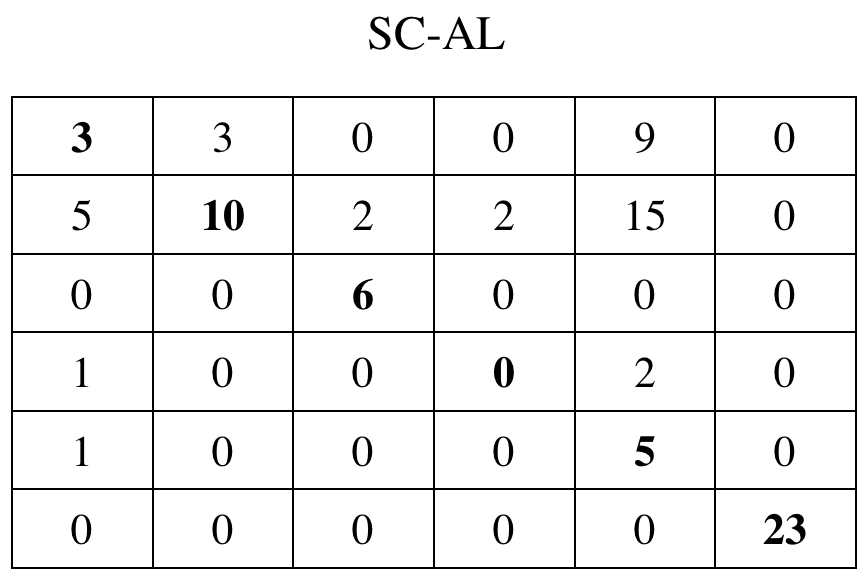}~\\
			\end{tabular}
	\end{center}
	\caption{Confusion matrices of seven combination methods on the MIT dataset}
	\label{fig:confusion}
\end{figure*}

Finally, in addition to the benchmarking results, the confusion matrices for different clustering methods with the MIT dataset are shown in Fig.~\ref{fig:confusion} as illustrative example of the clustering quality. The columns of the matrix represent the predicted clusters while the rows represent the intended classes. From the  diagonal entries of the matrices (which are the numbers of objects that have been correctly identified for each class), it is clear that \textbf{SC-SR} best reveals the 6 classes in the ground truth data.

\section{Related work}
In this section we give a review of the literature that is related to our work. We start with the general field of graph-based data processing and learning techniques. Next, we move onto spectral methods applied on graphs. Finally, we discuss a series of existing works that involves a framework of combining information from multiple graphs.

Nowadays, graph theory is widely considered as a powerful mathematical tool for data modeling and processing, especially when the pairwise relationships between objects are of interest. In practice, it is highly connected with a major branch of scientific research, that is, network analysis. Hence, graph-based data mining and analysis have become extremely popular over the last two decades. In \cite{Holder06} the authors have described the recent developments on the theoretical and practical aspects of the graph-based data mining problems together with a sample of practical applications. Especially, graph-based clustering  has attracted a large amount of interests due to its numerous applications. In \cite{Schaeffer07} the author has investigated the state-of-the-art techniques and recent advances in this vibrant field, from hierarchical clustering to graph cuts, spectral methods and Markov chain based methods. These are certainly the foundations of our work. From a methodology point of view, regularization theory on graphs is of particular interests. In \cite{Smola03}, the authors have developed the regularization theory of learning on graphs using the canonical family of kernels on graphs. In \cite{Zhou04}, the authors have defined a family of differential operators on graphs, and used them to study the ``smoothness" measure of the functions on graphs. They have then proposed a regularization framework based on this smoothness measure. These works provide the main inspirations that lead to our second approach.

In addition to the general graph-based data processing, there is a unique branch in graph theory that is devoted to analyzing the spectrum of the graphs, the spectral graph theory. The manuscript by Chung \cite{Chung97} gives a good introduction to this field. Among various methods that are developed, we particularly emphasize the so-called spectral clustering algorithm, which has become one of the major graph-based clustering techniques. Due to its promising performance and close links to other well-studied mathematical fields, a large number of variants of the original algorithm has been proposed, such as the constrained spectral clustering algorithm \cite{Xu05}\cite{Lu08}\cite{Li09}\cite{Li09_2}\cite{Wang10}. In general, these works have suggested different ways to incorporate constraints in the clustering task. Among them, \cite{Li09} has proposed a regularization framework in the graph spectral domain, which provides the closest methodology to our work.

Recently, data that can be represented by multiple graphs has aroused increasing attention. In the literature of learning community this is often referred to as ``multiple views" or ``multiple kernels", which intuitively means we investigate data from different viewpoints. In this setting, the general problem is how to efficiently combine information from multiple graphs for our analyses. In this sense, the following research efforts have the closest ideas to our presented work. In \cite{Argyriou05}, the authors have proposed a method to compute an ``optimal combined kernel" for combining graphs. Their idea is essentially based on averaging the graph Laplacian matrices. In \cite{Zhou07}, the authors has modeled spectral clustering on a single graph as a random walk process, and then proposed a mixed random walk when two graphs are given. However, the way they make the combination is still based on a convex combination of the two graphs. In \cite{Muthukrishnan10}, the authors have presented a novel way to exploit the relationships between different graph layers, which permits efficient combination of multiple graphs by a regularization framework in the signal domain. In \cite{Cheng09}, the authors have proposed to achieve the final clustering result by post-processing the result from each individual graph layer. In \cite{Savas10} and \cite{Vasuki10}, the authors have worked with very similar settings to our work, however the problems they have tackled there are not clustering. Finally, the work by Tang in \cite{Tang09} is the closest to our first algorithm \textbf{SC-GED} in the sense that they also use a unified matrix factorization framework to find a joint low dimensional representation shared by the multiple graphs, which directly enlightened us to develop our first approach. Very recently, Kumar \cite{Kumar10} proposed the co-regularization framework which is conceptually similar to our second algorithm \textbf{SC-SR}, and is adopted as a competing method in our experiments.

To summarize, although some of the works mentioned above are closely related to what we have presented in this paper, there are still noticeable differences that can be summarized as follows. First, despite the nature of the spectral clustering algorithm, most of the existing efforts to combine information from multiple graph layers are done in the signal domain, while the well-developed spectral techniques are mostly applied on a single graph. In contrast, our proposed methods provide novel ways to do the same task in the graph spectral domain. Second, to the best of our knowledge, in almost all the state-of-the-art algorithms for combining multiple graphs, different graph layers are either treated equally or combined through a weighted sum. However, we propose \textbf{SC-SR} based on a spectral regularization process, in which individual graph layers play different roles in the combination process. In addition, we suggest to quantitatively measure the respective importance of different graph layers from an information-theoretic point of view, which could be beneficial for processing multiple graphs in general. Third, there are only a few works that address the problem of clustering with multiple graph layers, especially in the context of mobile social network analysis. We believe that our efforts to work with rich mobile phone datasets are good attempts in this emerging field.

\section{Conclusion}
In this paper we study the problem of clustering with data that can be represented by a multi-layer graph. We have shown that generalizations of the well-developed spectral techniques applied on a single graph are of great potential in such emerging tasks. In particular, we have proposed two novel methodologies to find a joint spectrum that is shared by all the graph layers: a joint matrix factorization approach and a graph-based spectral regularization framework. In the second approach, we suggest to treat individual graph layers based on their respective importances, which are measured through an information-theoretic point of view. In addition to the improvements we get in the clustering benchmarks with three social network datasets, we believe that the concept of joint spectrum shared by multiple graphs is of broad interest in graph-based data processing tasks, as it suggests one way to generalize the classical spectral analysis to multi-dimensional cases. This is certainly one of the focuses in our future work.

\section{Acknowledgement}
This work has been partly funded by Nokia Research Center (NRC), Lausanne, Switzerland. The authors would like to thank Prof. Radu Horaud and Prof. Nathan Eagle for the permission of using their figures in this paper (Fig.~\ref{fig:multigraph} and Fig.~\ref{fig:mapping}). The authors are also grateful to Prof. Weinman for the kind sharing of the MATLAB implementation of the L-BFGS algorithm\footnotemark[6].

\footnotetext[6]{\url{http://www.cs.grinnell.edu/~weinman/code/index.shtml}}

\bibliographystyle{IEEEtran.bst}
\bibliography{mybibfile.bib}

\begin{thebibliography}{10}
\providecommand{\url}[1]{#1}
\csname url@samestyle\endcsname
\providecommand{\newblock}{\relax}
\providecommand{\bibinfo}[2]{#2}
\providecommand{\BIBentrySTDinterwordspacing}{\spaceskip=0pt\relax}
\providecommand{\BIBentryALTinterwordstretchfactor}{4}
\providecommand{\BIBentryALTinterwordspacing}{\spaceskip=\fontdimen2\font plus
\BIBentryALTinterwordstretchfactor\fontdimen3\font minus
  \fontdimen4\font\relax}
\providecommand{\BIBforeignlanguage}[2]{{%
\expandafter\ifx\csname l@#1\endcsname\relax
\typeout{** WARNING: IEEEtran.bst: No hyphenation pattern has been}%
\typeout{** loaded for the language `#1'. Using the pattern for}%
\typeout{** the default language instead.}%
\else
\language=\csname l@#1\endcsname
\fi
#2}}
\providecommand{\BIBdecl}{\relax}
\BIBdecl

\bibitem{Kumar10}
A.~Kumar, P.~Rai, and {H. Daum{\'e} III}, ``{Co-regularized Spectral Clustering
  with Multiple Kernels},'' \emph{NIPS 2010 Workshop: New Directions in
  Multiple Kernel Learning}, 2010.

\bibitem{Schaeffer07}
E.~Schaeffer, ``{Graph clustering},'' \emph{Computer Science Review}, vol.~1,
  no.~1, pp. {27--64}, 2007.

\bibitem{Eagle10}
N.~Eagle, A.~Clauset, A.~Pentland, and D.~Lazer, ``{Multi-Dimensional Edge
  Inference},'' \emph{Proceedings of the National Academy of Sciences (PNAS)},
  vol. 107, no.~9, p. {E31}, 2010.

\bibitem{Eagle06}
N.~Eagle and A.~Pentland, ``{Reality Mining: Sensing Complex Social Systems},''
  \emph{Personal and Ubiquitous Computing}, vol.~10, no.~4, pp. {255--268},
  2006.

\bibitem{Shi00}
J.~Shi and J.~Malik, ``{Normalized Cuts and Image Segmentation},'' \emph{IEEE
  Trans. Pattern Anal. and Mach. Intell.}, vol.~22, no.~8, pp. 888--905, Aug
  2000.

\bibitem{Eagle09}
N.~Eagle, A.~Pentland, and D.~Lazer, ``{Inferring Social Network Structure
  Using Mobile Phone Data},'' in \emph{Proceedings of the National Academy of
  Sciences}, vol. 106, no.~36, 2009, pp. {15\,274--15\,278}.

\bibitem{Luxburg07}
U.~von Luxburg, ``{A Tutorial on Spectral Clustering},'' \emph{Statistics and
  Computing}, vol.~17, no.~4, pp. {395--416}, 12 2007.

\bibitem{MacQueen67}
J.~B. MacQueen, ``{Some Methods for classification and Analysis of Multivariate
  Observations},'' \emph{Proceedings of 5th Berkeley Symposium on Mathematical
  Statistics and Probability}, p. pp. 281Ð297, 2010.

\bibitem{Lovasz96}
L.~Lov{\'a}sz, ``{Random Walks on Graphs: A Survey},'' \emph{Combinatorics,
  Paul Erd{\"o}s is Eighty}, vol.~2, pp. {353--398}, J{\'a}nos Bolyai
  Mathematical Society, Budapest, 1996.

\bibitem{Stewart90}
G.~W. Stewart and J.~Sun, ``{Matrix Perturbation Theory},'' \emph{Academic
  Press, New York}, 1990.

\bibitem{Bhatia97}
R.~Bhatia, ``{Matrix Analysis},'' \emph{Springer, New York}, 1997.

\bibitem{Nocedal06}
J.~Nocedal and S.~J. Wright, ``{Numerical Optimization},'' \emph{Springer
  Series in Operations Research and Financial Engineering}, 2006.

\bibitem{Tang09}
W.~Tang, Z.~Lu, and I.~Dhillon, ``{Clustering with Multiple Graphs},'' in
  \emph{International Conference on Data Mining}, Miami, Florida, USA, Dec
  2009.

\bibitem{Chung97}
F.~Chung, ``{Spectral Graph Theory},'' \emph{American Mathematical Society},
  1997.

\bibitem{Horaud}
R.~Horaud, ``{A Short Tutorial on Graph Laplacians, Laplacian Embedding, and
  Spectral Clustering}.''

\bibitem{KSMH09}
\BIBentryALTinterwordspacing
D.~Knossow, A.~Sharma, D.~Mateus, and R.~P. Horaud, ``Inexact matching of large
  and sparse graphs using laplacian eigenvectors,'' in \emph{Proceedings 7th
  Workshop on Graph-based Representations in Pattern Recognition}, ser. LNCS
  5534.\hskip 1em plus 0.5em minus 0.4em\relax Venice, Italy: Springer, May
  2009. [Online]. Available:
  \url{http://perception.inrialpes.fr/Publications/2009/KSMH09}
\BIBentrySTDinterwordspacing

\bibitem{SHKV09a}
\BIBentryALTinterwordspacing
A.~Sharma, R.~P. Horaud, D.~Knossow, and E.~von Lavante, ``Mesh segmentation
  using laplacian eigenvectors and gaussian mixtures,'' in \emph{Proceedings of
  AAAI Fall Symposium on Manifold Learning and its Applications}, ser. Fall
  Symposium Series Technical Reports.\hskip 1em plus 0.5em minus 0.4em\relax
  Arlington, VA: AAAI Press, November 2009. [Online]. Available:
  \url{http://perception.inrialpes.fr/Publications/2009/SHKV09a}
\BIBentrySTDinterwordspacing

\bibitem{Zhou04}
D.~Zhou and B.~Sch{\"o}lkopf, ``{A Regularization Framework for Learning from
  Graph Data},'' in \emph{ICML Workshop on Statistical Relational Learning and
  Its Connections to Other Fields}, 2004, pp. 132--137.

\bibitem{Zhou04-2}
D.~Zhou, O.~Bousquet, T.~Lal, J.~Weston, and B.~Sch{\"o}lkopf, ``{Learning with
  Local and Global Consistency},'' in \emph{Advances in Neural Information
  Processing Systems (NIPS)}, vol.~16, Cambridge, MA, 2004, pp. 321--328.
  (Eds.) S. Thrun, L. Saul and B. Sch{\"o}lkopf, MIT Press.

\bibitem{Zhu03}
X.~Zhu, Z.~Ghahramani, and J.~Lafferty, ``{Semi-supervised learning using
  Gaussian fields and harmonic functions},'' in \emph{The 20th International
  Conference on Machine Learning (ICML)}, 2003.

\bibitem{Bousquet04}
O.~Bousquet, O.~Chapelle, and M.~Hein, ``{Semi-supervised learning using
  Gaussian fields and harmonic functions},'' in \emph{Advances in Neural
  Information Processing Systems (NIPS)}, vol.~16, Cambridge, MA, 2004, pp.
  (Eds.) S. Thrun, L. Saul and B. Sch{\"o}lkopf, MIT Press.

\bibitem{Zhou04-3}
D.~Zhou, J.~Weston, A.~Gretton, O.~Bousquet, and B.~Sch{\"o}lkopf, ``{Ranking
  on Data Manifolds},'' in \emph{Advances in Neural Information Processing
  Systems (NIPS)}, vol.~16, Cambridge, MA, 2004, pp. 169--176. (Eds.) S. Thrun,
  L. Saul and B. Sch{\"o}lkopf, MIT Press.

\bibitem{deSa05}
V.~R. de~Sa, ``{Spectral Clustering with two views},'' in \emph{Proceedings of
  the Workshop on Learning with Multiple Views, ICML}, 2005.

\bibitem{Kiukkonen10}
N.~Kiukkonen, J.~Blom, O.~Dousse, D.~Gatica-Perez, and J.~Laurila, ``{Towards
  Rich Mobile Phone Datasets: Lausanne Data Collection Campaign},'' in
  \emph{International Conference on Pervasive Services}, Berlin, Germany, Jul
  2010.

\bibitem{Newman02}
M.~Girvan and M.~E.~J. Newman, ``{Community Structure in Social and Biological
  Networks},'' \emph{Proceedings of the National Academy of Sciences (PNAS)},
  vol.~99, pp. 7821--7826, 2002.

\bibitem{Newman04}
M.~E.~J. Newman and M.~Girvan, ``{Finding and evaluating community structure in
  networks},'' \emph{Phys. Rev. E}, vol.~69, p. 026113, 2004.

\bibitem{Newman042}
M.~E.~J. Newman, ``{Fast algorithm for detecting community structure in
  networks},'' \emph{Phys. Rev. E}, vol.~69, p. 066133, 2004.

\bibitem{Nefedov11}
N.~Nefedov, ``{Multi-Membership Communities Detection in Mobile Networks},'' in
  \emph{Workshop of the International Conference on Web Intelligence, Mining
  and Semantics}, May 2011.

\bibitem{Blondel08}
V.~D. Blondel, J.-L. Guillaume, R.~Lambiotte, and E.~Lefebvre, ``{Fast
  unfolding of communites in large networks},'' \emph{Journal of Statistical
  Mechanics: Theory and Experiment}, vol. 1742-5468, no. P10008, p. (12 pp.),
  2008.

\bibitem{Manning08}
C.~D. Manning, P.~Raghavan, and H.~Sch{\"u}tze, ``{Introduction to Information
  Retrieval},'' \emph{Cambridge University Press}, 2008.

\bibitem{Holder06}
D.~Cook and L.~Holder, ``{Mining Graph Data},'' 2006.

\bibitem{Smola03}
A.~Smola and R.~Kondor, ``{Kernels and Regularization on Graphs},'' in
  \emph{16th Annual Conference on Computational Learning Theory}, Washington,
  DC, USA, Aug 2003.

\bibitem{Xu05}
Q.~Xu, M.~desJardins, and K.~Wagstaff, ``{Constrained Spectral Clustering under
  a Local Proximity Structure},'' in \emph{Proceedings of the 18th
  International Florida Artificial Intelligence Research Society (FLAIRS)
  Conference}, May 2005.

\bibitem{Lu08}
Z.~Lu and M.~{\'A}. Carreira-Perpi{\~n}{\'a}n, ``{Constrained spectral
  clustering through affinity propagation},'' in \emph{IEEE Conf. Computer
  Vision and Pattern Recognition}, 2008.

\bibitem{Li09}
Z.~Li, J.~Liu, and X.~Tang, ``{Constrained clustering via spectral
  regularization},'' in \emph{IEEE Conf. Computer Vision and Pattern
  Recognition}, 2009.

\bibitem{Li09_2}
Z.~Li and J.~Liu, ``{Constrained Clustering by Spectral Kernel Learning},'' in
  \emph{Proc. IEEE Int'l Conf. Computer Vision (ICCV)}, 2009.

\bibitem{Wang10}
X.~Wang and I.~Davidson, ``{Flexible constrained spectral clustering},'' in
  \emph{ACM SIGKDD Conference on Knowledge Discovery and Data Mining},
  Washington, DC, Jul 2010, pp. 563--572.

\bibitem{Argyriou05}
A.~Argyriou, M.~Herbster, and M.~Pontil, ``{Combining Graph Laplacians for
  Semi-Supervised Learning},'' in \emph{NIPS}, 2005.

\bibitem{Zhou07}
D.~Zhou and C.~Burges, ``{Spectral Clustering and Transductive Learning with
  Multiple Views},'' in \emph{Proceedings of the 24th International Conference
  on Machine Learning (ICML)}, 2007, pp. 1159--1166.

\bibitem{Muthukrishnan10}
P.~Muthukrishnan, D.~Radev, and Q.~Mei, ``{Edge Weight Regularization Over
  Multiple Graphs For Similarity Learning},'' in \emph{IEEE International
  Conference on Data Mining}, Sydney, Dec 2010.

\bibitem{Cheng09}
Y.~Cheng and R.~Zhao, ``{Multiview spectral clustering via ensemble},'' in
  \emph{IEEE International Conference on Granular Computing}, Nanchang, Aug
  2009, pp. 101--106.

\bibitem{Savas10}
B.~Savas, W.~Tang, Z.~Lu, and I.~S. Dhillon, ``{Supervised Link Prediction
  Using Multiple Sources},'' in \emph{IEEE International Conference on Data
  Mining}, Sydney, Dec 2010.

\bibitem{Vasuki10}
V.~Vasuki, N.~Natarajan, Z.~Lu, and I.~S. Dhillon, ``{Affiliation
  Recommendation using Auxiliary Networks},'' in \emph{Proceedings of the 4th
  ACM Conference on Recommender Systems(RecSys)}, Sep 2010.

\end{thebibliography}

\end{document}